\documentclass[5p]{elsarticle}
\usepackage[utf8]{inputenc}
\usepackage{hyperref}
\usepackage{graphicx}
\usepackage{caption}
\usepackage{subcaption}
\usepackage{lipsum}
\usepackage{amsmath}
\usepackage{setspace}
\usepackage{xcolor}
\newcommand{\revtext}[1]{#1}

\title{Real-time simulation of viscoelastic tissue behavior with \\physics-guided deep learning}

\date{\today}

\begin{document}

\begin{frontmatter}
\author[1,2]{Mohammad Karami}\corref{cor1}
\ead{mohammad.karami@mirametrix.com}
\cortext[cor1]{Corresponding Author: Mohammad Karami}
\author[2]{Hervé Lombaert}
\ead{herve.lombaert@etsmtl.ca}

\author[1]{David Rivest-Hénault}
\ead{david.rivest-henault@nrc-cnrc.gc.ca}

\address[1]{National Research Council of Canada, 75 de Mortagne Blvd, Boucherville, Québec, Canada}
\address[2]{ETS Montreal, Department of Computer and Software Engineering, Canada}

\begin{abstract}
  Finite element methods (FEM) are popular approaches for simulation of soft tissues with elastic or viscoelastic behavior. However, their usage in real-time applications, such as in virtual reality surgical training, is limited by computational cost. In this application scenario, which typically involves transportable simulators, the computing hardware severely constrains the size or the level of details of the simulated scene. To address this limitation, data-driven approaches have been suggested to simulate mechanical deformations by learning the mapping rules from FEM generated datasets. Prior data-driven approaches have ignored the physical laws of the underlying engineering problem and have consequently been restricted to simulation cases of simple hyperelastic materials where the temporal variations were effectively ignored. However, most surgical training scenarios require more complex hyperelastic models to deal with the viscoelastic properties of tissues. This type of material exhibits both viscous and elastic behaviors when subjected to external force, requiring the implementation of time-dependant state variables. Herein, we propose a deep learning method for predicting displacement fields of soft tissues with viscoelastic properties. \revtext{The main contribution of this work is the use of a physics-guided loss function for the optimization of the deep learning model parameters.}
 The proposed deep learning model is based on convolutional (CNN) and recurrent layers (LSTM) to predict spatiotemporal variations. It is augmented with a mass conservation law in the lost function to prevent the generation of physically inconsistent results. The deep learning model is trained on a set of FEM datasets that are generated from a commercially available state-of-the-art numerical neurosurgery simulator. The use of the physics-guided loss function in a deep learning model has led to a better generalization in the prediction of deformations in unseen simulation cases. Moreover, the proposed method achieves a better accuracy over the conventional CNN models, where improvements were observed in unseen tissue from 8\% to 30\% depending on the magnitude of external forces. \revtext{It is hoped that the present investigation will help in filling the gap in applying deep learning in virtual reality simulators, hence improving their computational performance (compared to FEM simulations) and ultimately their usefulness.}
\end{abstract}


\end{frontmatter}

\section{Introduction}
Neurosurgical training is typically obtained in operating rooms under the direct supervision of experienced neurosurgeons, who are severely limited both in available time and in the number of opportunities to practice. \revtext{Consequently, extraclinical surgical simulation is regularly integrated in training curriculum to provide for more development opportunities. Evidences from the literature tend to demonstrate that specific skills acquired during simulation translate to the operating room \cite{Badash}. Traditionally, surgical simulation use live animal, bench, or cadaveric models \cite{Davies}.} Recently, virtual reality simulators have become a promising alternative method to meet educational requirements \cite{Delorme, Brunozzi_2018, Vikal, Law, Badash}. \revtext{The use of simulation for surgical training can have a measurable and significant clinical impact by enabling new possibilities in surgical training. In a randomized control trial, \cite{Fazlollahi} demonstrated that using a simulator with metric-based artificial intelligence (AI) tutoring is more effective than receiving remote tutoring by a human instructor, or no tutoring while learning simulated brain tumor resections. In this case, it is the realistic simulation system with bimanual haptic feedback that provides the monitoring data that is required by the AI agent, therefore enabling this application. Additionally, the availability of high precision data streams from surgical simulation allows to automatically categorize the level of expertise of the trainee using machine learning algorithms \cite{WinklerSchwartz}. Continuous expertise monitoring systems can further assess surgical bimanual performance in real-time, which could be leveraged to provide predictive validation during surgical residency training, allowing the early detection of errors \cite{Yilmaz} and more efficient training. Virtual reality simulators thus enable trainees to practice on a variety of educational scenarios as well as to enable the definition of new training metrics and applications (e.g. remote training).}

\subsection{Overview of virtual reality simulators}
To provide realistic touch sensations through haptic feedback devices, virtual reality simulators require a real-time simulation of tissue deformation \cite{Bo}. The behavior of the tissue is critical during neurosurgery and its simulation is challenging due to its complex nonlinear viscoelastic nature. Finite Element Methods (FEM) are often used numerical techniques for this type of simulation. However, the application of FEM in real-time simulation is currently limited by hardware requirements. \revtext{While several types of procedure have been successfully simulated, for example, Cranial Microneurosurgery  \cite{Delorme}, others, which require a larger field of view or involve stiffer non-rigid material, are much more challenging to implement. Examples of challenging scenarios include femoral nailing in orthopedic \cite{Luca} or endonasal neurosurgical procedures. If a simulated scene is compute-bound, the designer needs to fall-back on ad-hoc heuristic methods, which might degrade realism.}

At the technical level, one of the main challenges in FEM is a concurrent update of matrix entries, describing the physical state, by several computational threads. This is difficult to implement efficiently in parallel due to the memory access arrangement \cite{Ljungkvist}. The following body of work has been proposed to overcome this limitation.

A first type of solutions are those based on model-order reductions \cite{Niroomandi_2008, Niroomandi_2012, Taylor_2011}. The conventional method for constructing a model with comparatively fewer degrees of freedom is the Proper Orthogonal Decomposition (POD). POD extracts the main displacement patterns, known as modes, that constitute simplified representations of the tissue displacements. Then a linear combination of these modes is used to approximate the final displacement field. The use of POD for real-time simulation of tissue behavior has been extensively studied \cite{Yvonnet_2007, Hernandez, Kerfriden} and with satisfying results in linear problems. However, this approach fails in the simulation of highly nonlinear problems \cite{Bhattacharjee, Chatterjee, Niroomandi_2013}, such as in neurosurgery simulations. This limitation was circumvented by expressing the modes as a sum of separable nonlinear functions, referred to as Proper Generalized Decomposition (PGD) \cite{Niroomandi_2013}. However, due to the implementation of boundary conditions as a new dimension into PGD, this technique is not well generalized and needs to be modified and combined with kernel principal component analysis \cite{Gonzalez}.
    
A second alternative for accelerating FEM simulations exploits machine learning approaches \cite{Phellan}. Based on FEM simulation data, machine learning models learn a function that maps an input, such as external forces, to an output, such as nodal displacements, without any previous knowledge of the problem. It has been postulated that if a machine learning model is trained on a FEM dataset, it can be subsequently used to predict the outcome of FEM simulations in real time. Such use of machine learning for real-time simulation of tissue behavior has been applied to simulate various organs, including breasts \cite{Martinez} and livers \cite{Oscar, Lorente}. However in these works the involved machine learning techniques, notably Support Vector Machine (SVM) and random forest, restrict their application to small nodal displacements, as discussed by \cite{Mendizabal}.
   
Recently, Convolutional Neural Networks (CNNs), a class of machine learning methods, have been used to speed up FEM simulations\cite{Mendizabal, Pfeiffer, Han}. CNNs are shown to be powerful in extracting complex patterns with the advantage of parallelizing well on modern GPUs. Mendizabal et al. \cite{Mendizabal} have used CNNs, based on a U-Net architecture \cite{Ronneberger}, for predicting nodal displacements of an elastic liver in real-time. They have also compared the CNN model with the POD method \cite{Goury} and have shown that the accuracy of a CNN is higher than that of a POD in a comparable inference time. Pfeifer et al. \cite{Pfeiffer} have also used fully convolutional neural networks for predicting nodal displacements of three different liver models and have obtained a reasonable accuracy within 20~ms of inference time. \revtext{Han el al. \cite{Han} used a CNN model in conjunction with a classical solver to approach a multiphysics problem in electrosurgery.}
   
Despite the wide use of machine learning in computer vision and natural language processing, applications to problems in physics have often met limited success \cite{Karpatne, Rai}. This issue is partly due to the absence of physical laws in machine learning algorithms. For instance, Jia et al. \cite{Jia} have highlighted this issue in the prediction of lake temperature profiles. The calculation of the lake temperature is rather similar to the calculation of material displacement (our case) since both variables can be obtained from physical principles of momentum/energy conservation. Jia et al. \cite{Jia} showed that the predicted temperature profiles from deep learning techniques are not consistent with respect to the energy equation if the physical law of the problem is absent in the loss function. For this reason, previous deep learning techniques (CNNs) \cite{Mendizabal, Pfeiffer} that have suggested speeding up FEM simulations were restricted to simulation cases of simple hyperelastic materials where the temporal variations were effectively ignored. Combining physical laws with deep learning, referred to as physics-guided deep learning, has received considerable attention in recent years. Willard et al. \cite{Willard} have provided a comprehensive review on this subject.

\subsection{Contribution}
  \revtext{There is a substantial difference in the effectiveness of virtual reality simulators achieved using machine learning (data-driven) based systems, compared with FEM-simulation systems utilizing only numerical methods. The main benefit of the former consists of their usage in real-time applications while providing similar results to FEM. Therefore, we addressed this problem in the research reported in this paper. This work proposes, for the first time to the authors' knowledge, a deep learning method for predicting displacement fields of soft tissues with viscoelastic properties.} 
  
  \revtext{Overall, our contribution consists of the integration of a mass-conservation law as a new domain-specific loss function into the proposed deep learning model, which makes it possible to prevent the production of physically inconsistent results. This is in contrast to the alternative existing machine learning techniques, which are focused exclusively on sample data, and they fail in the simulation of highly nonlinear problems like viscoelastic tissue behavior. To show the real application of the proposed method (as part of another contribution), we have used the real data from a virtual reality neurosurgery simulator, NeuroTouch, which is a commercial product. The data is unique and challenging because of the hyperelasticity behavior of the tissue. We have also demonstrated the faster computation of our proposed method in comparison to the virtual reality simulator results, which are based on FEM technique. It is hoped that the present investigation may also help in filling the gap in applying deep learning in FEM simulations.}

\subsection{Paper structure}
\revtext{The paper is organized as follows. In Section 2, we present some information about commercial neurosurgical simulators, with particular attention given to the implemented FEM technique in NeuroTech. The proposed model is described in Section 3, and the results of the proposed model are reported and discussed in Sections 4 and 5. The paper is concluded with Section 6.}

\section{Background information}
Commercial neurosurgical simulators are based on tissue deformation models derived from the fundamentals of continuum mechanics. In such real-time applications, a common strategy is to use simplified numerical FEM models that are less costly to compute than more analytical alternatives. Since one of the motivations for this work is to emulate the output of a FEM-based simulator using a data-driven approach (hopefully at a greater computational rate), the main driving equations are presented here. The selected FEM reference software, which will be presented in Sec.~\ref{sec:results}, uses an explicit time-integration scheme, and tissues are modeled as viscoelastic solids using a quasilinear viscoelastic constitutive model\footnote{In mechanics of materials, a constitutive model is a representation of the constitutive laws which are the governing equations describing the behavior of the material.}. The relaxation function is as follows:
 
\begin{equation} \label{eq:relax}
G(t) = g_0 + \sum_{k=1}^{2} g_k e^{-t/\tau_k}
\end{equation}

where $\tau_k$ is a relaxation time and $g_k$ is a relaxation modulus. The elastic part of tissue behavior is modeled as hyperelastic solids of the generalized Neo-Hookean constitutive model defined by the following strain energy density function:

\begin{equation} \label{eq:strain}
W = C_1 (I_1 - 3 -2\mathrm{ln}(j)) + \frac{K}{2}(j-1)^2,
\end{equation}
where $C_1$ is the material constant determined from experiments, $I_1$ is the first invariant of the left Cauchy-Green deformation tensor, $K$ is the bulk modulus, and $j$ is the square root of the third invariant of the left Cauchy-Green deformation tensor and a measure of the volumetric deformation. The bulk modulus is defined as follows:

\begin{equation} \label{eq:modulus}
K = \frac{2\mu(1+\nu)}{3(1-2\nu)},
\end{equation}
where $\mu$ is the shear modulus, and $\nu$ is the Poisson ratio. 

\section{Method}
In this section, data-driven approaches, based on deep neural network architectures, are proposed for speeding up the FEM simulation software. The deep learning models are trained on a set of FEM datasets generated from a commercial-grade simulator. The input to the model is an external force on each node, a tensor of $N_x\times N_y\times N_z\times 3$, where $N$ is the number of nodes in the three directions of $x,y,z$ (as an example, a brain tumor grid mesh used in this work for real-time simulation has size $N_x=17$, $N_y=17$, and $N_z=8$). The model output is a predicted nodal displacement field with the same shape as the input tensor.
  
\subsection{Existing Neural Network models for tissue displacement simulation}
U-Net, a neural network architecture based on convolutional layers, \cite{Ronneberger} has already been experimented successfully for the simulation of simple hyperelastic materials \cite{Mendizabal}. However, this model is only able to extract spatial characteristics of the nodal displacement field and ignores the temporal deformation of the viscoelastic material, which might reduce temporal smoothness and precision. In this work, we propose to alleviate these issues by defining a model that includes an LSTM layer. In addition, terms that are specific to the mechanical nature of the problem are also considered.

\subsection{Proposed model}
\subsubsection{Network architecture}
\begin{figure*}[t] 
    \centering
    \includegraphics[width=16cm,height=45mm]{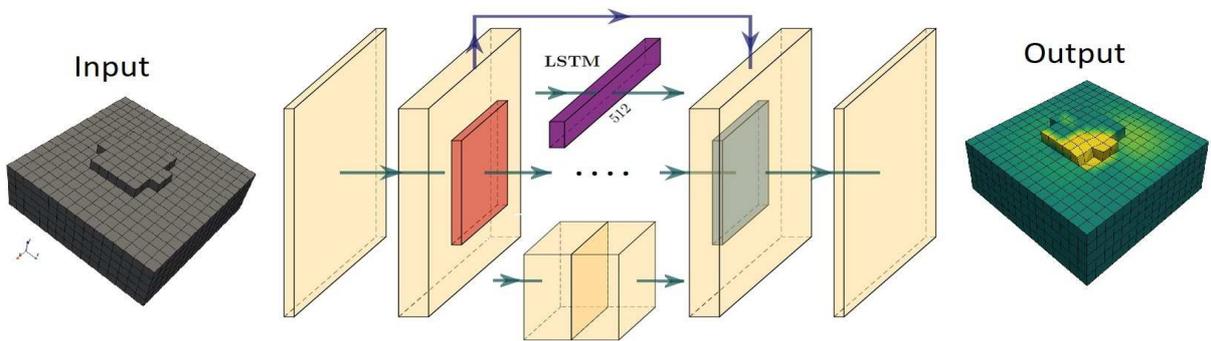}
    \caption{CNN U-Net and CNN-LSTM deep learning architectures.  The reference CNN U-Net network (see section~\ref{sec:unet}) does not include the purple LSTM box. The proposed CNN-LSTM general architecture is similar to that of the CNN network, except for the bottleneck (center layers) which consists of an LSTM layer. \revtext{The main contribution herein is the use of a physics-guided loss function for the optimization of the model parameters.}}
    \label{fig2}
\end{figure*}

Our proposed deep neural network model combines Convolutional Neural Network (CNN) and Long Short-Term Memory (LSTM) within the architecture shown in Figure~\ref{fig2}. LSTM, which is one of the Recurrent Neural Network (RNN) models, is different from CNN models in its consideration of time sequences. LSTM contains an input gate, an output gate, and a forget gate. The integration of these gates makes LSTM a suitable model for time-dependent problems. However, viscoelastic materials with complex hyperelastic models are very nonlinear to be modeled with a generic CNN-LSTM model. Herein, the CNN-LSTM model with a Mean Squared Error (MSE) loss function is referred to as "generic CNN-LSTM" model. This model is based on the CNN model, whose bottleneck consists of an LSTM layer with 512 nodes. Due to the power of the LSTM in feature extraction, we were able to reduce the number of convolutional operations in the model. The encoding/decoding path has a single convolutional operation with 32 filters. \revtext{The designed parameters of the network architecture are shown in Table \ref{ModelParameters}. It shows the settings and the number of trainable parameters of the entire layers, including convolutional, activation, pooling, LSTM, and upsampling layers. The model optimizer is Adam with a learning rate of 0.00001.}

\begin{table*} 
\caption{\label{ModelParameters} \revtext{The proposed CNN-LSTM architecture. Note that the merging layer consists of a concatenation of the outputs of two predecessor convolutional layers along the channel dimension}.}
\small
\begin{center}
\begin{tabular}{|c|c|c|c|c|c|} 
\hline
Type & \#Filters & Kernel size & Stride & UpSampling size & \# Parameters \\ 
\hline
Convolution & 32 & (3, 3, 3) & (1, 1, 1) & - & 2624\\ 
\hline
Activation (ReLu) & - & - & - & - & -\\ 
\hline
Pooling & - & (3, 3, 3) & (3, 3, 3) & - & -\\ 
\hline
Bi-LSTM(512) & - & - & - & - & 11538432\\ 
\hline
Activation (tanh) & - & - & - & - & -\\ 
\hline
UpSampling & - & - & - & (4, 4, 3) & -\\ 
\hline
Convolution & 32 & (3, 3, 3) & (1, 1, 1) & - & 27680\\ 
\hline
Activation (ReLu) & - & - & - & - & -\\ 
\hline
Merging* & - & - & - & - & -\\ 
\hline
Convolution & 3 & (1, 1, 1) & (1, 1, 1) & - & 195\\  
\hline
\end{tabular}
\end{center}
\end{table*}

\subsubsection{Mechanical displacement-specific loss function}
To improve the modeling of viscoelastic materials, we regularize the deep learning model by an auxiliary task to ensure that the nodal displacement maps are constrained by the constitutive laws. The conventional loss function in regression problems is an MSE between the model predictions, $\hat{U}_{n,t}$, and the FEM simulations output, $U_{n,t}$, for every node and every time step. 

To favorably constrain the learning process, a term based on mass conservation law can be considered to extend the loss function. When the model satisfies the mass conservation law, the volume of the tumor at each time step, $V_t$ should always remain constant and equal to the initial volume of undeformed material, $V_{origin}$. 
Thus, the predicted nodal displacement fields that violate, $\Delta V= V_t - V_{origin} = 0$, will be penalized. One usual way of penalization is using Rectified Linear Units (ReLU) as activation functions, but, in this study, it was observed that this function suffers from the vanishing gradient problem. Thus, to avoid the vanishing gradient inherent from ReLU, a squared difference between $V_t$ and $V_{origin}$, i.e., L2 loss function, has been considered for penalization. Finally, the combined loss function, referred to as physics-guided, is defined as follows:

\begin{equation}
\label{eqn_physics}
\begin{aligned}
Loss = L_{MSE} + \lambda L_{Physics} = \frac{1}{N} \sum_{t,n=1}^{T,N}  (\hat{U}_{n,t} - U_{n,t})^2 +\\ \lambda (V_t - V_{origin})^2 ,
 \end{aligned}
\end{equation}
where $\lambda$ is a hyper-parameter that regulates the balance between $L_{MSE}$ and $L_{Physics}$, and it is equal to 0.1. It was observed that decreasing $\lambda$ to 0.01 will cause the effects of $L_{Physics}$ to be small, or increasing $\lambda$ to 1 will deteriorate the learning process. It should be noted that the current FEM simulation has a coarse mesh with compressibility, leading us to have some cases with physical error. To circumvent this issue, the physics-guided term, $L_{Physics}$, has been applied only to the cases whose volume change is above a threshold value, which is 7\% of the material volume. In the rest of this paper, the CNN-LSTM model with the physics-guided custom loss function is referred to as our model.

It is worth mentioning that we also tried a different physical law for implementation into the physical loss function, $L_{Physics}$. Based on Hooke's law, the direction of the displacement at each node should be consistent with the direction of the external force at each node. In other words, the cosine angle, $\cos \theta$, between the external force vector of each node and the displacement vector of each node should be always zero. Thus, we added a new term, $ReLU(1-\cos \theta)$, into the $L_{Physics}$. However, it was found out this new term does not have a significant effect on the learning process, and thus it has been ignored in the proposed physics-guided loss function.

Once the deep learning model has been trained, it can be deployed for real-time simulation of tissue behavior.

\section{Results} \label{sec:results}
\begin{figure}[th]
        \centering
        \includegraphics[width=6cm,height=8cm]{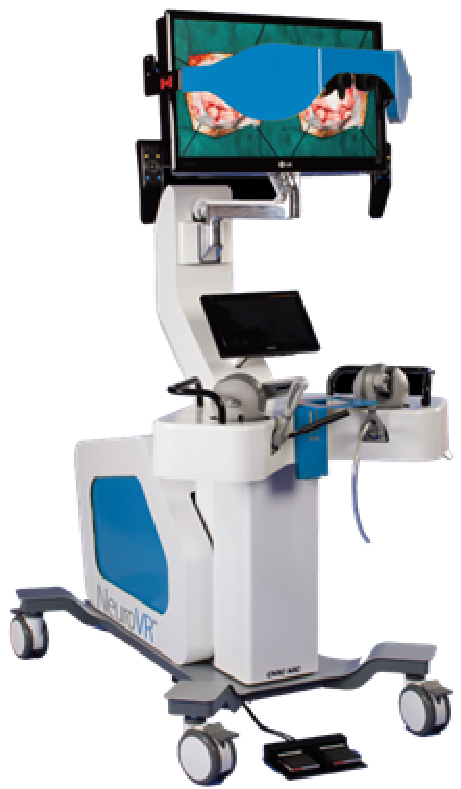}
        \caption{The virtual reality neurosurgery simulator, NeuroTouch, at the National Research Council of Canada.}
        \label{NeuroVR}
    \end{figure}
  
A virtual reality neurosurgery simulator with haptic feedback\footnote{\textit{NeuroTouch}, from the National Research Council of Canada, commercially available under the \textit{NeuroVR} trademark from CAE HealthCare.} \cite{Delorme, Brunozzi_2018}, (Figure \ref{NeuroVR}) was used to illustrate the applicability of the proposed method for fast viscoelastic tissue displacement simulation. While capable of real-time simulation, the simulation software embedded in the simulator is computationally bounded, which prevents operation on more portable hardware, or, alternatively, the use of bigger mechanical meshes. This surgical simulator has been used for generating FEM simulation dataset of a brain tumor. The tumor mesh has overall dimension $N_x=17$, $N_y=17$, and $N_z=8$, and consists of 1782 nodes and 1314 hexahedron cells, which leads to an input tensor size of $(17, 17, 8, 3)$. It is subjected to different states of two contact forces. The bottom and the peripheral surfaces of the material are subjected to a zero-displacement boundary condition, while its top surfaces are left free. The simulator implements equations (\ref{eq:relax}--\ref{eq:modulus}). 
In \eqref{eq:relax}, $\tau_1$ and $\tau_2$ are respectively 330s and 11s, and $g_1$ and $g_2$ are respectively 0.12 and 0.8. In \eqref{eq:strain} and \eqref{eq:modulus}, $C_1$, $\nu$ and $\mu$, are respectively 0.0002 Mpa, 0.0004 Mpa, and 0.42. 

\subsection{Comparison with alternative methods}
Herein, our proposed technique has been compared with two different techniques for the prediction of tissue deformation from the external force: linear regression (as a baseline model) and a previously introduced Convolutional Neural Network (CNN) model. The models have been implemented in TensorFlow, and the source code is available at: \url{https://github.com/Mkarami3/UNet_FEM}. 

\subsubsection{Baseline model}
The baseline model is a linear regression model. The linear regression model predicts the nodal displacement field as a linear combination of the input force. This model is not expected to give appealing results but is included to serve as a conceptual lower bound. For complex and nonlinear problems, nonlinear approaches, such as deep learning, are usually required.

\subsubsection{CNN U-Net model}\label{sec:unet}
The selected reference convolutional neural network is based on a U-Net architecture and is represented in Figure~\ref{fig2}. The encoding path consists of two convolutional operations with 64 filters followed by a max-pooling operation which divides the input size into half. The bottleneck consists of a stack of two convolutional operations with 128 filters. Its decoding is done by upsampling followed by six convolutional operations with respectively 64, 64, 128, 64, and 3 filters. U-Net \cite{Ronneberger} has already been experimented successfully for the simulation of simple hyperelastic materials \cite{Mendizabal}. However, this model is able to extract only spatial characteristics of the nodal displacement field and ignores the temporal deformation of the viscoelastic material. In contrast, with the proposed CNN-LSTM model, these issues have been circumvented by implementing the LSTM layer into our proposed model and by using a physics-guided loss function.

\subsection{Experiments on hyper-parameters optimization}
In this section, we have performed multiple experiments on different configurations of the proposed model to find a combination of hyper-parameters that improves the model performance. These hyper-parameters are the number of neurons ($N_N$) and the number of time-steps ($N_t$) in the LSTM layer. The size of the dataset is 1512, which is splitted into a training set with a size of 1210 and a validation set with a size of 300. All the experiments have been done on a personal computer with an Intel Core i7 (3.40 GHz) and an Nvidia Geforce RTX 2080 with 11 GB of video RAM.

\begin{figure}[t]
    \centering
    \includegraphics[width=9cm]{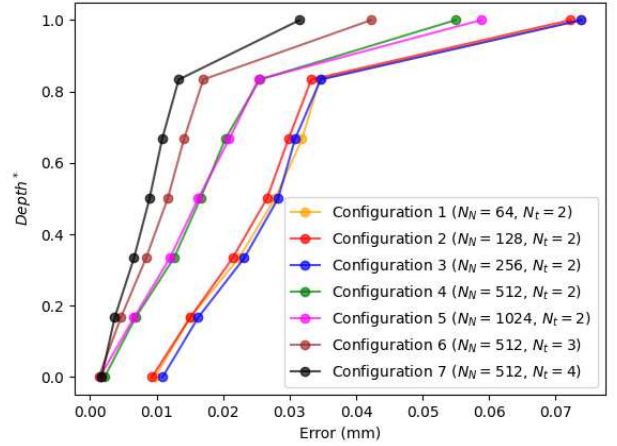}
    \caption{Variation of the average absolute error versus the normalized depth for the different configurations of the CNN-LSTM model (ours), see Table \ref{Experiments}.}
    \label{Exps}
\end{figure}

Table \ref{Experiments} presents the set of configurations of the LSTM layer, and Figure \ref{Exps} shows their corresponding performance in the prediction of the nodal displacement field. It shows that when the time-step ($N_t$) remains constant, increasing the number of neurons ($N_N$) from 64 to 512, results in a reduced displacement error at all ground truth displacement levels. This behavior is observable in Figure~3 by comparing the yellow line (configuration 1) with the green line (configuration 4). However, increasing $N_N$ value further, to 1024, does not have any effect on the reduction of the displacement error since the error of configuration 4 (green line) and the error of configuration 5 (pink line) collapse each other in Figure 3.

Briefly, this Figure shows that the proposed model with 512 neurons ($N_N$) and two time-steps ($N_t$) is optimal with respect to accuracy and real-time application. Note that while the model accuracy improves with the number of time-steps, herein, $N_t = 2$ is selected to meet the requirement of a real-time processing application.

\begin{table*} 
\caption{\label{Experiments} Different sets of configurations of the LSTM layer in our model.}
\small
\begin{center}
\begin{tabular}{|c|c|c|c|} 
\hline
& \#Neurons ($N_N$) & \#Time-steps ($N_t$) & \#Trainable parameters \\ 
\hline
Configuration 1 & 64 & 2 & 1,219,235 \\ 
\hline
Configuration 2 & 128 & 2 & 2,501,155\\ 
\hline
Configuration 3 & 256 & 2 & 5,261,603\\ 
\hline
Configuration 4 & 512 & 2 & 11,568,931\\ 
\hline
Configuration 5 & 1024 & 2 & 27,329,315\\ 
\hline
Configuration 6 & 512 & 3 & 11,568,931\\ 
\hline
Configuration 7 & 512 & 4 & 11,568,931\\ 
\hline
\end{tabular}
\end{center}
\end{table*}

\subsection{Assessment of the accuracy of the proposed model}

\begin{figure}[t]
    \centering
    \includegraphics[width=9cm]{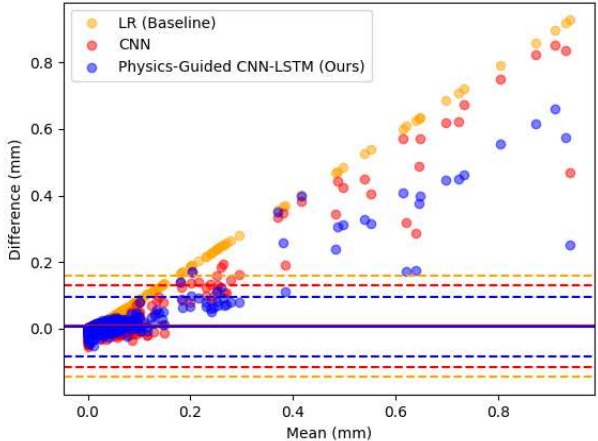}
    \caption{Plot of differences versus mean of the predicted deformations of the linear regression (baseline), the CNN, and the \revtext{Physics-Guided} CNN-LSTM (ours) models from the reference FEM dataset. The solid lines represent the mean of differences, and the dashed lines represent the 96\% confidence interval of the limits.}
    \label{fig4}
\end{figure}

To assess the performance of the proposed model, Bland-Altman plot is shown in Figure~\ref{fig4}. This analysis is performed by averaging the predicted nodal displacement fields over time sequences and then presenting the resulting graph as a scatter plot, in which the vertical axis shows the difference between the predicted nodal displacement and the reference FEM simulation dataset and the horizontal axis represents the mean of the reference FEM simulation dataset. This technique has advantages over the correlation coefficient and regression approaches in considering the differences between the performance of the different models.

The Bland-Altman plot (Fig.~\ref{fig4}) shows us that the linear regression model, as a baseline, is not able to predict the nodal displacement field at all. This observation is expected as the linear regression is based on an assumption that the nodal displacement field is a linear combination of the input force, which is not a true assumption here due to the viscoelasticity property of the material. While the \revtext{CNN} deep learning models look capable of modeling tissue deformation, our model has better accuracy than the CNN model when the material is subjected to high strain, see the differences between the two models when the mean of deformation is higher than 0.6 mm. Moreover, the differences do not lie in the 96\% confidence interval limit since they are skewed and are not normally distributed.

Figure~\ref{fig5} shows the performance of the proposed deep learning model for estimating the internal deformation of an organ. The plot shows the error variations in millimeters of the normalized depth, $depth^*$, of the material. The depth is normalized with the height of the material, and the displacement error is calculated by averaging $(\hat{U}_{n,t} - U_{n,t})$ in time sequences and the nodes which have a similar depth. It shows that the CNN model has an average of 45\% improvement over the regression model (as a baseline). By implementing the LSTM layer into the proposed model, this improvement can be increased further to an average of 31\% over the CNN model. 

\subsubsection{Contribution of the physics-guided term}
\begin{figure}
    \centering
    \includegraphics[width=8cm]{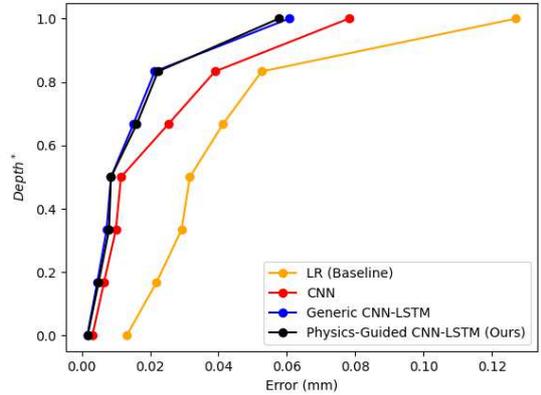}
    \caption{Variation of the average absolute error versus the normalized depth for the baseline, the CNN, the generic CNN-LSTM, and the physics-guided CNN-LSTM (ours) models.}
    \label{fig5}
\end{figure}

\begin{figure}
     \centering
     \begin{subfigure}{8cm}
         \centering
         \includegraphics[width=\textwidth]{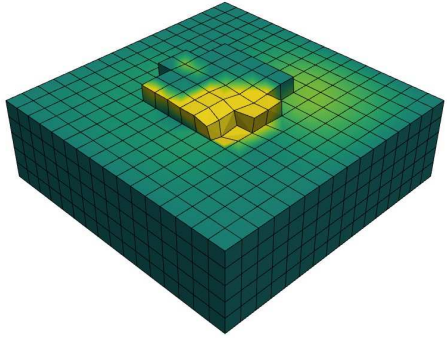}
         \caption{}
         \label{fig7a}
     \end{subfigure}
     \begin{subfigure}{8cm}
         \centering
         \includegraphics[width=\textwidth]{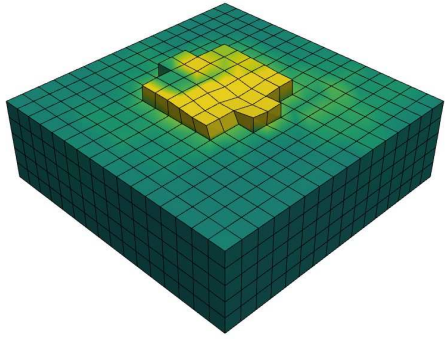}
         \caption{}
         \label{fig7b}
     \end{subfigure}
    \begin{subfigure}{8cm}
        \centering
        \includegraphics[width=\textwidth]{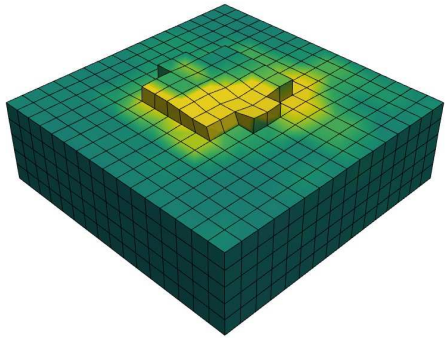}
        \caption{}
    \end{subfigure}
    \caption{Nodal displacement field of (a) the reference FEM simulation, (b) as predicted by the generic CNN-LSTM model, and (c) as predicted by the physics-guided CNN-LSTM model (ours). A supplementary video related to these plots can be found in the online version of the article.}
    \label{fig7}
\end{figure}

\begin{figure}[t]
     \centering
     \begin{subfigure}[p!]{9cm}
         \centering
         \includegraphics[width=\textwidth]{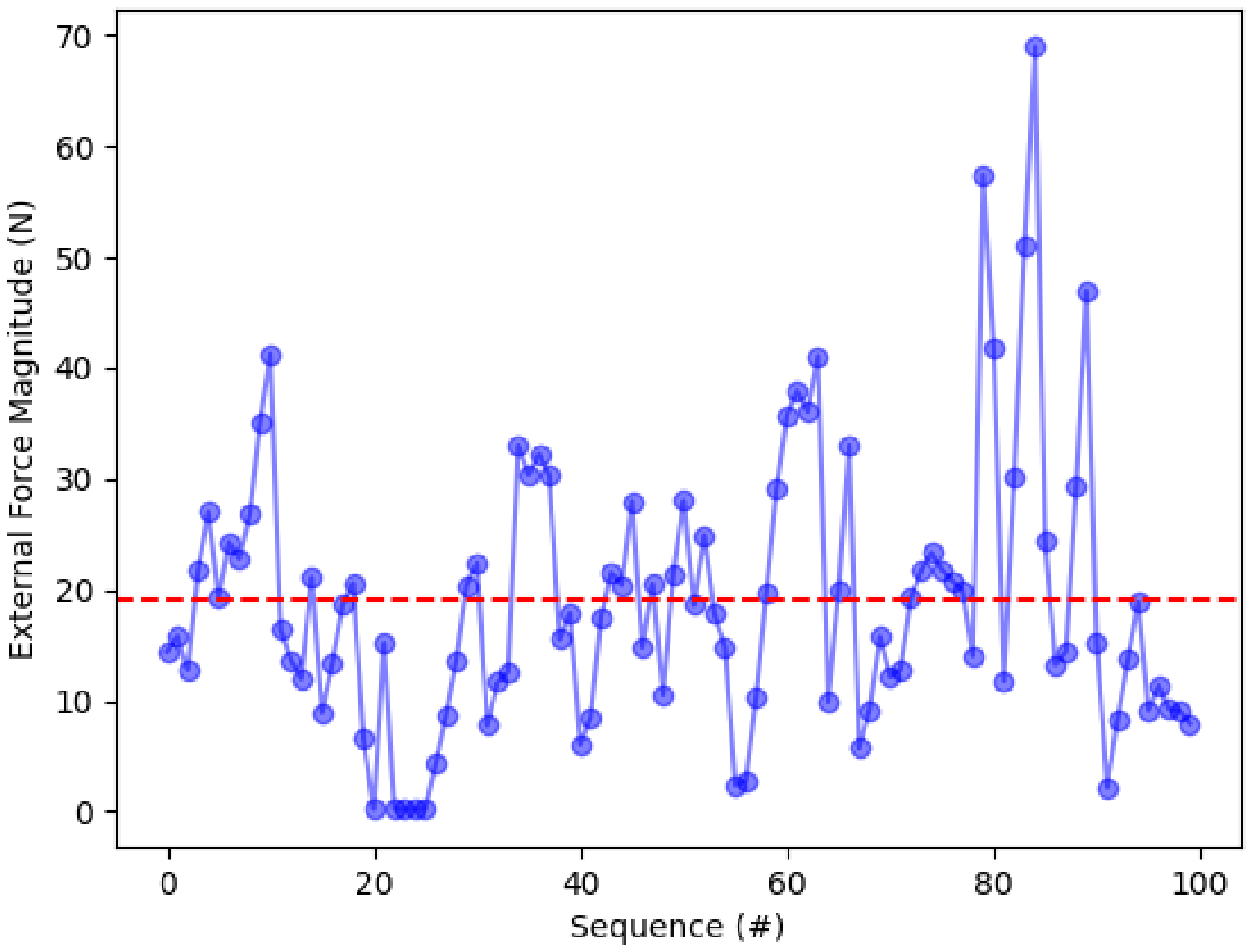}
         \caption{}
         \label{fig6a}
     \end{subfigure}
     \vfill
     \begin{subfigure}[p!]{9cm}
        \centering
        \includegraphics[width=\textwidth]{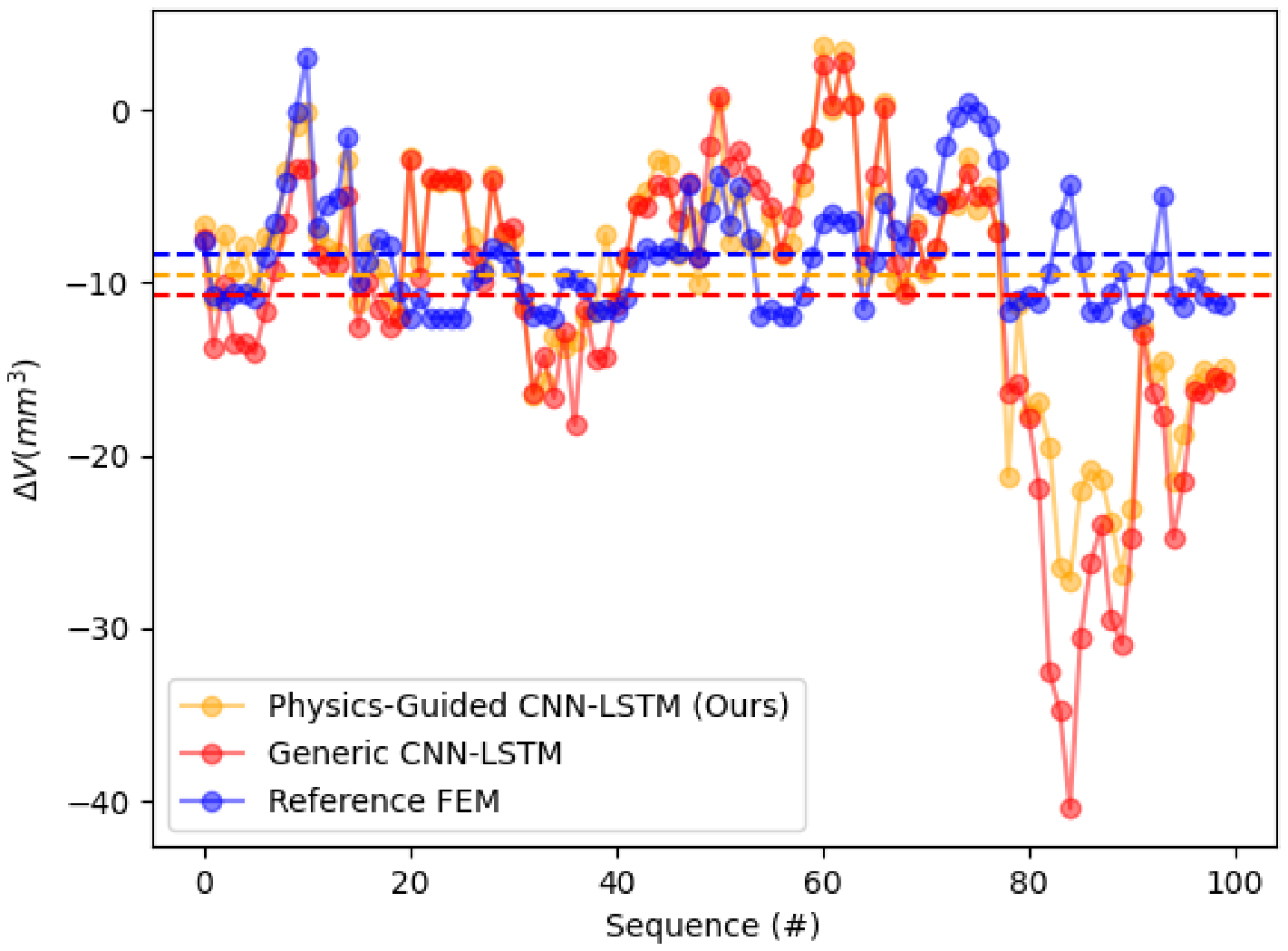}
         \caption{}
         \label{fig6b}
     \end{subfigure}
     \caption{Variations of (a) the external force magnitude, and (b) the amount of violation in volume preservation,$\Delta V(mm^3)$ over the 100 sequences of test cases in the reference FEM, generic and physics-guided CNN-LSTM (ours) models.}
     \label{fig6}

\end{figure}

The specific contribution of the proposed physics-guided term in the loss function has been assessed by comparing the proposed physics-guided CNN-LSTM with a CNN-LSTM network optimized using the L2 loss function, and with the CNN U-Net and linear baselines. The results are presented in Figure \ref{fig5}.
Even though nodal displacement prediction errors of the generic CNN-LSTM model and our proposed model are shown to be close. The main difference between these two models lies in the generalization of deep learning. The proposed model generalizes well on the unseen simulation cases through consideration of the mass-conservation law. Such an unseen FEM simulation case and its predicted displacement field are shown in Figure \ref{fig7}. This property has been also demonstrated by monitoring the amount of violation in volume preservation, $\Delta V(mm^3)$, and the external force magnitude over 100 sequences of the testing dataset, see Figure \ref{fig6}. We can see that the generic CNN-LSTM, in comparison to the proposed model, has a maximum violation in the sequence~83 for which the magnitude of the external force has a maximum deviation from its mean value. In other words, when the input data (external force) of a testing case is more deviated from what has been seen in the training dataset, a higher error in the generalization of the generic CNN-LSTM model will be observed. However, the physical loss function provides properties favorable for a better generalization in our proposed model. For unseen simulation cases (i.e., sequences 78 to 85 in Figure~\ref{fig6}), this improvement is in the range of 8\% to 30\%, depending on the magnitude of the external force.

It should be noted that for a perfect model that satisfies the mass conservation law, $\Delta V(mm^3)$ should always remain constant and equal to zero. However, the current FEM simulation has a small number of elements, resulting in a coarse mesh with some convergence error, and the material is not perfectly incompressible. Thus, a small violation in volume preservation can be observed in the reference FEM cases, and our model gives a penalty only to those cases for which the physical inconsistency is above a threshold, i.e., $\Delta V > 12.1\, mm^3$.


\subsection{Computational performance} \label{sec:deployment}
In this section, we observe the computational behavior of the proposed deep learning method and analyze if it could be used to speed up FEM simulations in real time. The proposed model was deployed on CPU and compared its inference time with the latency time of FEM simulation software of NeuroTouch. At the time of this study, it was not possible to deploy CNN-LSTM on the Nvidia GPU as the required LSTM component was not supported by TensorRT 7.2.3. To have an estimation of the inference time of the proposed model on GPU, we deployed the CNN model on GPU.

\begin{figure}[t]
    \centering
    \includegraphics[width=10cm]{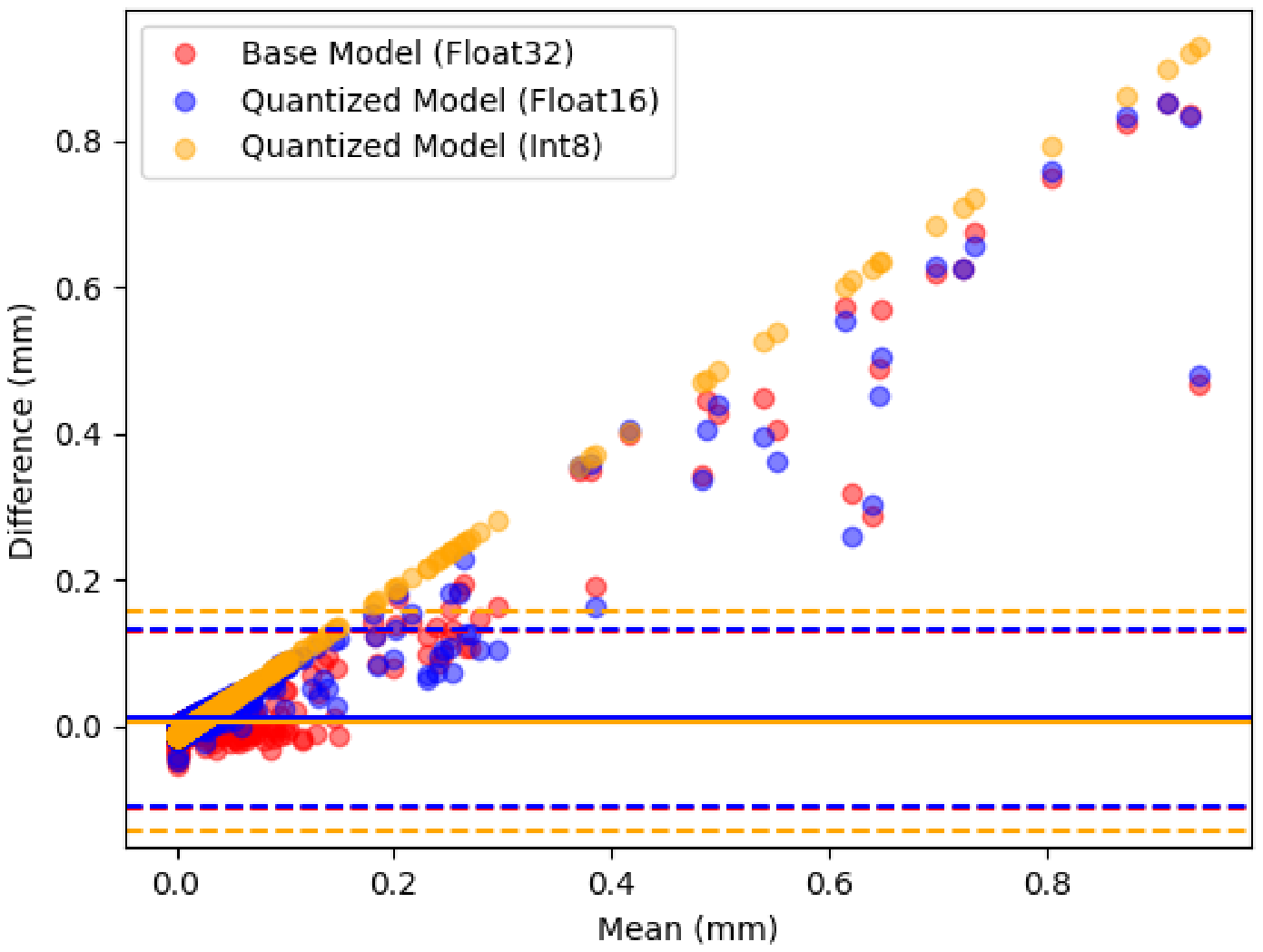}
    \caption{Effects of quantization on the CNN model accuracy. Quantization can speed up the model inference at the cost of losing accuracy.}
    \label{quantization}
\end{figure}

Table~\ref{inference_time} presents the inference time of the CNN and the proposed model in comparison to the latency time of NeuroTouch. It can be seen that the CNN deep learning model can be almost eleven times faster than NeuroTouch when it is deployed on the GPU. Moreover, it was found out that by pruning low magnitude weights in the convolutional layers of the CNN model, inference time can be 1.3 times faster on CPU without losing accuracy. This acceleration on CPU can be improved further by quantization operation at the cost of the model accuracy. Figure~\ref{quantization} shows the effects of quantization on the accuracy of the CNN model. While the effects of a float16 quantization on the model accuracy are not significant, the model loses its performance for int8 quantization. Deployment of the CNN model on GPU suggests that we can reduce the inference time of the proposed model and make it faster than NeuroTouch when it is deployed on GPU.

Currently, the proposed CNN-LSTM architecture cannot be deployed on the GPU. However, when that functionality becomes available, extrapolation based on CPU times suggests that it could possibly run 3 times faster than the explicit solver. Although the current implementation is slower than the NeuroTouch FEM software, there might still be benefits in its deployment on CPU. The implemented FEM engine in NeuroTouch uses an explicit solver with a latency fixed at 0.010s, which is a design parameter. The engine achieves a fixed latency by performing a fixed number of integration steps, which depends on the size of the mesh. Thus, for some configurations, particularly in the case of large deformations on larger mesh, the explicit solver might not fully converge to the ideal solution. Therefore, it is up to the designer to provide a mesh that is sufficiently small to be solvable with a precision that is acceptable for the simulation most of the time. However, this issue does not exist in the deployment of the deep learning models, and thus, herein, it is believed that the advantages of our model architecture in deployment outweigh the disadvantages of a slower latency time.

\begin{table}
\caption{\label{inference_time} Comparison of inference time between the deep learning models and the NeuroTouch FEM software. Note that the FEM computation time is fixed at 0.010, but the convergence of the solver is not guaranteed.}
\begin{center}
{\footnotesize
\begin{tabular}{|c|c|c|} 
\hline
Model  & Latency on CPU (s) & Latency on GPU (s)\\ 
\hline
CNN & 0.022 & 0.0019 \\ 
\hline
Proposed model  & 0.032 & Not Supported\\ 
\hline
NeuroTouch-FEM & Not Real-Time & $0.01^*$\\ 
\hline
\end{tabular}}\\[2mm]
\end{center}
\end{table}

\section{Discussion}  
Our experiments have demonstrated that physics-guided deep learning is a viable solution for speeding up FEM simulations. However, there are still limitations in this approach which can be important challenges for future work. 

The first limitation is that the physics-guided loss function, $L_{Physics}$ in Eq.~\ref{eqn_physics}, is only valid for FEM cases that do not have a large amount of violation in volume preservation, $\Delta V=0$. If the reference FEM simulation cases have a large convergence error (resulting from a coarse mesh) or exhibit a high compressibility value, then $\Delta V$ would have a large deviation from zero, which makes the present $L_{Physics}$ invalid.

The second limitation is that the capacity of the LSTM model for capturing temporal dependencies is restricted to viscoelastic materials that exhibit short stress relaxation. Viscoelastic materials have both viscous and elastic properties when subjected to deformations. Herein, the stress relaxation depends on the factor $\tau$, and the higher its value, the longer it takes for the stress to relax. This requires an LSTM layer with larger input time steps and a greater number of neurons, which is not feasible for usage in real-time processing applications.

Another limitation is that the present deep learning network is not able to model FEM simulation cases that include topology changes. Virtual cutting of deformable objects, or cauterization, requires remeshing and movement of nodes in the cutting layer. However, the present network requires a mesh-like tensor with a fixed node order to apply convolution operation.
  
\section{Conclusions}
In this paper, we propose a deep learning method for the prediction of the displacement field of soft tissues with viscoelastic properties. \revtext{The main contribution of this work is the use of a physics-guided loss function for optimization of the deep learning model parameters}. To deal with the viscoelastic property of the material with a complex hyperelastic model, the proposed deep learning model is augmented with a constitutive law, referred to as physics-guided deep learning (implemented as a new loss function) as well as an LSTM layer. These two features enable the deep learning model to predict both spatial and temporal variations of the nodal displacement field. It was found that the proposed method achieves a better accuracy over the CNN model and has a better generalization over the generic model with MSE loss function. In terms of computational cost, it was shown that the proposed method is currently capable of running in real-time on the CPU (0.032s/cycle, 31Hz), but at a relatively low rate. However, a potential GPU implementation, when the required software components become available, is expected to bring a speedup of about 10x, which would make this implementation 3x faster than the reference explicit solver. Meanwhile, there are other benefits in deploying CNN-LSTM on CPU as the reference NeuroTouch FEM engine uses an explicit solver which can have convergence problems that are not present in deep learning models.

\section*{Acknowledgments}
This work has been partially supported by the National Research Council Canada (NRC), the Natural Sciences and Engineering Research Council of Canada (NSERC), as well as computational resources from Compute Canada and Calcul Quebec.

\end{document}